\documentclass[runningheads]{llncs}
\usepackage[T1]{fontenc}
\usepackage{graphicx}
\usepackage{booktabs}
\usepackage[misc]{ifsym}

\usepackage{amssymb} 
\usepackage{amsmath} 
\usepackage{bm}      
\usepackage{graphicx} 
\usepackage{algorithm}
\usepackage{algpseudocode}
\usepackage{xcolor}

\usepackage{tabularx}
\usepackage[hidelinks]{hyperref}
\usepackage{mwe}

\begin{document}

\renewcommand{\floatpagefraction}{0.8} 

\title{Hybrid-IR: Dual-Path Hybrid Retrieval with Iterative Reasoning for Complex Medical Question Answering}

\titlerunning{Hybrid Retrieval with Iterative Reasoning for Complex Medical QA}

\toctitle{Hybrid-IR: Dual-Path Hybrid Retrieval with Iterative Reasoning for Complex Medical Question Answering}

\author{Sheng Wan \inst{1} \and
Jiahui Zhang\inst{1} \and
Zicheng Zhao\inst{2}\and
Shougang Ren \inst{1} \Letter}

\authorrunning{Sheng Wan et al.}
\tocauthor{Sheng Wan, Jiahui Zhang, Zicheng Zhao, Shougang Ren}

\institute{College of Artificial Intelligence, Nanjing Agricultural University, Nanjing, China \email{wansheng315@hotmail.com, zhangjiah@stu.njau.edu.cn, rensg@njau.edu.cn}
\and
School of Computer Science and Engineering, Nanjing University of Science and Technology, Nanjing, China \email{zicheng.zhao@njust.edu.cn}
}

\maketitle              

\begin{abstract}
Large language models (LLMs) have shown promising performance across a wide range of biomedical applications, including medical question answering (QA), yet they remain prone to hallucinations and outdated knowledge. Although retrieval-augmented generation (RAG) can alleviate this issue by incorporating external documents, there still exist two fundamental limitations. First, medical knowledge is often fragmented across documents, while most RAG methods rely on a single retrieval path, which makes it challenging to jointly preserve fine-grained semantic information and structured global associations. Second, static retrieval strategies are typically insufficient to support deep reasoning that is important in complex medical QA. In this paper, we present a dual-path retrieval framework with an iterative retrieval–reasoning mechanism termed ``Hybrid-IR'' for complex medical QA. The proposed Hybrid-IR integrates graph-based retrieval for exploration of structured knowledge and dense retrieval for fine-grained semantic matching. Moreover, the reasoning trajectory can be progressively refined through an iterative retrieve-reason loop. Experiments on three widely used medical QA benchmarks demonstrate the effectiveness of our Hybrid-IR.

\end{abstract}

\section{Introduction}
Large language models (LLMs)  \cite{openai2024hellogpt4o,yang2025qwen3} have demonstrated significant potential across a wide range of biomedical applications, particularly in medical question answering (QA) tasks that require deep reasoning over specialized domain knowledge \cite{singhal2025toward,sun2025cortexdebate,sun2025fast}. Despite their strong performance, deploying LLMs in complex medical scenarios remains challenging. One major limitation is that LLMs are prone to generating hallucinations that appear plausible yet lack factual basis \cite{hadi2024evaluation}. In addition, the reliance of LLMs on static pre-training data restricts their ability to incorporate the latest medical knowledge and clinical guidelines \cite{luo2025gfm}. As a result, the reliability of LLM-based medical QA systems remains inadequate.

Retrieval-augmented generation (RAG) \cite{lewis2020retrieval} offers an effective solution to reduce hallucinations by integrating external documents into the generation process and providing strict generation constraints, which enables LLMs to answer questions with up-to-date knowledge. However, applying RAG to the biomedical domain presents unique challenges. One key challenge is that existing RAG methods typically rely on a single retrieval path, either vector-based or graph-based, which makes it difficult to simultaneously preserve fine-grained semantic information and structured global associations \cite{cheng2025survey}. Vector-based retrieval methods typically encode documents as isolated embeddings that capture semantic similarity but fail to explicitly model the structured relationships among biomedical entities \cite{luo2025gfm,karpukhin2020denseretrieval}. This fragmented knowledge representation severely constrains the ability to synthesize cross-document insights, particularly when evidence is distributed across diverse sources in the biomedical domain. Differently, graph-based retrieval methods (\textit{e.g.}, GraphRAG \cite{edge2024graphrag}) introduce structured graphs to model dependencies among entities, which enhances the ability to capture global contextual relationships \cite{peng2025graph,Wan2026NeRS}. However, the graph construction process often fails to preserve fine-grained textual semantics, which may reduce the precision of evidence grounded in original medical documents. In addition to this limitation, most existing methods rely on static retrieval strategies that perform document retrieval only once before the generation process \cite{xiong2024improving}. Such design is often insufficient for complex biomedical questions that require multi-step reasoning. Intermediate reasoning steps often reveal additional information needs, yet the decoupling between retrieval and reasoning prevents the system from dynamically refining its retrieval focus. Consequently, static retrieval may fail to obtain the evidence required for subsequent reasoning steps, and thus resulting in incomplete evidence chains \cite{trivedi2023interleaving,xiong2024improving}.

To address these challenges, we propose a new method termed ``Hybrid-IR'' that enhances RAG for biomedical QA through a dual-path iterative reasoning-retrieval scheme. Specifically, we design a KG indexing (KG-index) module that constructs a structured cross-document index from extracted triples, which explicitly links dispersed medical entities and relations across documents. In addition, a document layer is introduced to align entity nodes with their source document chunks through provenance relations, which enables seamless transitions from abstract entity-level reasoning to grounded textual evidence. In parallel, dense retrieval is devised to capture unstructured fine-grained semantic details within the original text, which compensates for information that may be lost during graph construction. To enable deep and comprehensive reasoning for complex biomedical questions, we further integrate dual-path retrieval into an iterative framework, which enables a continuous ``retrieve–reason–retrieve'' loop. This mechanism enables cumulative evidence synthesis and dynamic refinement of reasoning trajectories across reasoning steps. By leveraging knowledge acquired in reasoning rounds to guide subsequent retrieval, the system effectively optimizes the reasoning path toward target solutions, thereby overcoming the limitations of static evidence aggregation in complex multi-hop tasks. We evaluate the proposed Hybrid-IR using three closed-book medical QA benchmarks, where no oracle documents are provided. Experimental results show that Hybrid-IR improves the average accuracy of strong backbone LLMs, including Llama-3-8B-Instruct \cite{llama3modelcard} and GPT-4o-mini \cite{openai2024hellogpt4o}, by up to 10 percentage points. The contributions are summarized as follows:
\begin{itemize}
  \item We propose a dual-path retrieval framework that jointly captures fine-grained semantic information and structured relational knowledge, which enables comprehensive evidence aggregation for complex medical QA.
  \item We design an iterative reasoning-retrieval mechanism that progressively accumulates evidence and dynamically refines reasoning trajectories through multiple rounds of ``retrieval-reasoning-retrieval''.
  \item Extensive experiments across three medical QA benchmarks demonstrate the effectiveness of Hybrid-IR, which achieves consistent performance gains across both open-source and commercial LLM backbones.
\end{itemize}

\section{Related Work}
\subsection{Retrieval-Augmented Generation}
RAG integrates external knowledge into LLMs to facilitate generation by retrieving relevant documents. Early works typically rely on pre-trained dense embedding models to encode chunked documents as discrete vectors, followed by retrieval and reranking based on similarity between queries and documents (\textit{e.g.} Contriever, Colbert) \cite{izacard2021contriever,khattab2020colbert}. Despite their efficiency and generalizability, treating documents as independent retrieval units makes it difficult to capture the underlying semantic relationships between them. Recent methods begin to introduce structured graphs to explicitly model knowledge relations. Specifically, distinct graph construction paradigms have emerged to model knowledge at different levels of granularity\cite{luo2025graph,zhao2023towards,zhao2025graph}. For instance, early methods like KGP \cite{KGP} use hyperlinks and KNN similarity, but miss semantic associations. RAPTOR \cite{raptor} builds hierarchical trees via recursive summarization. GraphRAG \cite{edge2024graphrag} uses LLMs to extract entities and relations, forming hierarchical graphs with community detection and summarization. HippoRAG \cite{jimenez2024hipporag} and HippoRAG 2 \cite{hipporag2} apply OpenIE to induce KGs that capture factual relationships. Building upon these foundations, advanced approaches such as G-reasoner \cite{luo2025g} propose refined multi-layered graph construction strategies, organizing knowledge into four hierarchical levels (\textit{e.g.}, entity, cluster, category, and document levels) to manage logical dependencies \cite{Wan2026Contrastive}. Beyond structural construction, the effective utilization of these graph-structured sources is crucial for bridging the gap between structural context and retrieval precision. HippoRAG and HippoRAG 2 employ a personalized PageRank algorithm on graphs to identify relevant knowledge, thereby strengthening multi-hop reasoning capability. Similarly, GFM-RAG \cite{luo2025gfm} introduces a graph foundation model based on Graph Neural Networks (GNNs \cite{wu2020comprehensive_gnn,zhao2023towards,zhao2025graph}), which enables robust reasoning across diverse datasets.

\subsection{Retrieval-Augmented Generation for Medical Question Answering}

Biomedical QA poses additional challenges due to specialized terminology and the need to integrate evidence scattered across multiple scientific documents. Researchers have proposed several retrieval methods to enhance biomedical information retrieval and RAG frameworks. For instance, MedCPT \cite{jin2023medcpt} is a dense embedding retriever and reranker specifically customized for the biomedical field. It addresses the scarcity of query-article annotation data in the biomedical domain by leveraging large-scale user click logs from PubMed. Beyond domain-specific retrieval models, several RAG frameworks have been developed for biomedical QA. The MedRAG toolkit \cite{xiong2024benchmarking} provides a comprehensive benchmark for evaluating RAG systems in the biomedical domain. It adopts a retrieval strategy that combines sparse and dense retrieval to extract relevant documents from the MedCorp corpus. Similarly, MedGraphRAG \cite{wu2025medical} constructs triple-based KGs to model intricate relationships among knowledge entities and employs U-retrieval techniques to enhance evidence retrieval efficiency. To address multi-step reasoning requirements in biomedical QA, i-MedRAG \cite{xiong2024improving} progressively expands evidence through an iterative retrieval mechanism, enabling the model to gather additional context for multi-step QA tasks.
However, most existing methods rely primarily on a single retrieval paradigm. Although MedRAG combines sparse and dense retrievers, it does not explicitly exploit structured relationships among documents. Conversely, MedGraphRAG focuses on structured graph evidence but may overlook fine-grained textual semantics. Furthermore, although iterative retrieval has been explored (e.g., i-MedRAG), it typically operates as a standalone evidence expansion loop decoupled from the LLM's intermediate reasoning states. To cope with issues, we propose a dual-path retrieval framework with an iterative retrieval–reasoning mechanism for complex biomedical QA. Unlike existing RAG frameworks that rely on a single retrieval paradigm, Hybrid-IR integrates the two paradigms within an iterative retrieval-reasoning mechanism, which enables reasoning feedback to dynamically refine hybrid evidence retrieval.

\section{Methodology}

\begin{figure}[!t]
	\centerline{\includegraphics[width=0.9\textwidth]{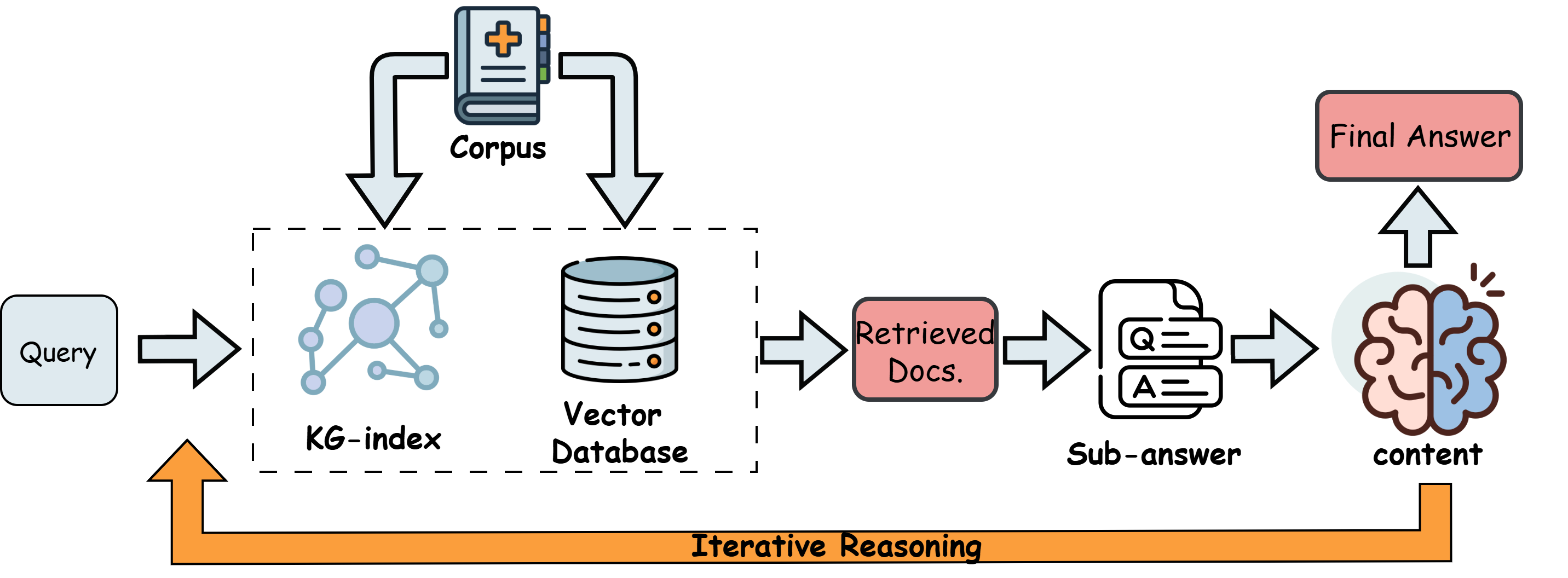}}
	\vskip -5pt
	\caption{The overall architecture of the proposed Hybrid-IR framework.}
\label{fig:1}
\end{figure}

This section details our proposed Hybrid-IR (see Fig.~\ref{fig:1}). Specifically, we illustrate the critical components of Hybrid-IR by explaining the offline hybrid index construction (see Section~\ref{offlineHybrid}), presenting online dual-path retrieval (see Section~\ref{3.2}), and elaborating iterative retrieval-reasoning for QA (see Section~\ref{IterativeQA}).

\subsection{Offline Hybrid Index Construction}
\label{offlineHybrid}

KGs explicitly model dependencies among entities, enabling structured cross-document indexing across multiple sources. Motivated by these advantages, we construct the knowledge-graph-based index to explicitly organize and connect dispersed medical evidence across documents. Given a document collection $\mathcal{D} = \{ d_1, d_2, \ldots, d_n \}$, we first perform medical named entity recognition and relation extraction on the documents to extract entities $\mathcal{E}$ and relations $\mathcal{R}_{\text{ent}}$, forming triples $\mathcal{T}\subseteq \mathcal{E}\times \mathcal{R}_{\text{ent}}\times \mathcal{E}$. Such a process can be achieved by existing open information extraction (OpenIE) tools \cite{openie,openie2}. These triples form the initial knowledge layer of the KG-index. To mitigate homonymy and semantic ambiguity in medical texts, we use the Contriever-MSMARCO \cite{izacard2021contriever} to conduct the entity resolution by computing pairwise cosine similarity between entity representations. Undirected synonymy relations are introduced between entity pairs whose similarity exceeds a threshold $\tau$, resulting in an augmented triple set $\mathcal{T}^{+}$. This augmentation enriches the knowledge layer by connecting highly similar entities that may appear across different documents. We then build an entity-to-document inverted index $\mathbf{M} \in \{0, 1\}^{|\mathcal{E}| \times |\mathcal{D}|}$, where $\mathbf{M}_{ij}=1$ indicates that entity $e_i \in \mathcal{E}$ is mentioned in document $d_j$, and $0$ otherwise. To enable explicit provenance tracking of evidence, each entity is linked to the document chunk from which it is extracted, forming a document layer parallel to the knowledge layer. In addition, we introduce a document layer to capture inter-document relations (\textit{e.g.}, content similarity or citation links). Specifically, we construct a heterogeneous KG, 
$G = G_{\text{ent}} \cup G_{\text{link}} \cup G_{\text{doc}}$. 
The entity-level subgraph $G_{\text{ent}}$ consists of entity–entity relational triples derived from the original and augmented triple sets,
\begin{align}
    G_{\text{ent}} &= \{ (e_i, r_{\text{ent}}, e_j) \mid (e_i, r_{\text{ent}}, e_j) \in \mathcal{T} \cup \mathcal{T}^{+} \},
\end{align}
while the entity–document linking subgraph $G_{\text{link}}$ is constructed based on the entity-to-document inverted index,
\begin{align}
    G_{\text{link}} &= \{ (e_i, r_{\text{link}}, d_k) \mid \mathbf{M}_{ik}=1 \}.
\end{align}
In addition, $G_{\text{doc}}$ encodes inter-document relations, where an edge is added between two documents if they are related by content similarity or citation links. Content similarity edges are constructed based on cosine similarity between document embeddings. Further implementation details and statistical characteristics of the KG-index construction are provided in the supplementary material.

\begin{figure}[!t] 
    \centering
    \includegraphics[width=0.99\textwidth]{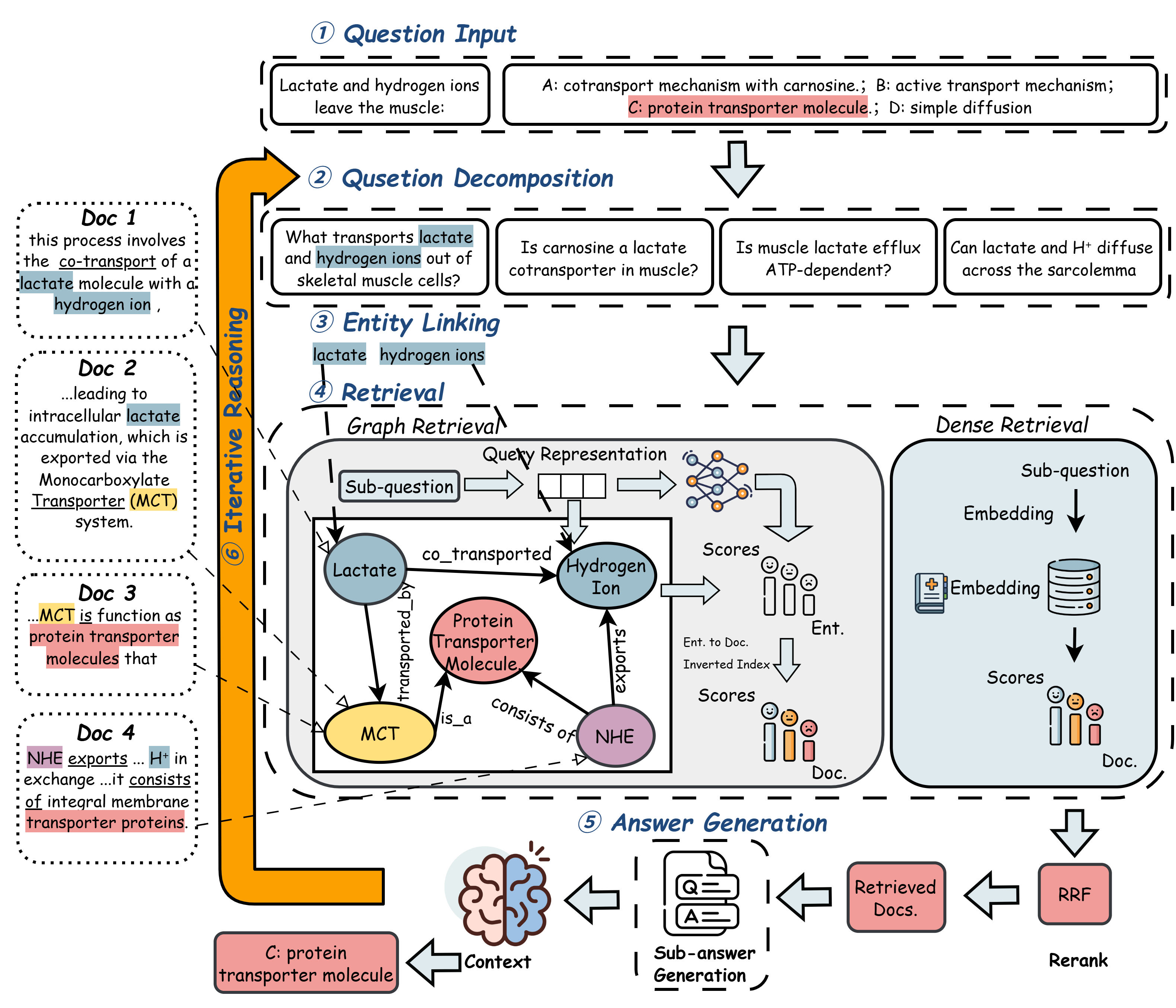}
    \caption{Overview of online dual-path retrieval and iterative retrieval-reasoning. Given a complex medical question, the model first decomposes it into a set of sub-questions. For each sub-question, evidence is retrieved in parallel via graph retrieval, which explicitly models dependencies among entities, and Dense Retrieval, which captures fine-grained semantic information from unstructured text. 
The retrieved results from both paths are then fused using Reciprocal Rank Fusion (RRF) \cite{rrf} to generate sub-answers, which are incorporated into the context to guide subsequent reasoning iterations.}
    \label{fig:2}
\end{figure}

In parallel with the KG indexing, vector indexing serves as a semantic complement that preserves fine-grained textual evidence which is difficult to fully represent with structured triples. 
Here, we encode each document $d_i$ into a dense vector $\mathbf{d}_{i}=\phi(d_i)$ using a pre-trained embedding model $\phi(\cdot)$ (\textit{i.e.}, MedCPT \cite{jin2023medcpt}), and store all vectors in a vector database for efficient similarity search.

\subsection{Online Dual-Path Retrieval}
\label{3.2}

Based on the offline-constructed KG and vector index, the framework performs online retrieval through a dual-path mechanism, as illustrated in Fig.~\ref{fig:2}. The key intuition behind this design is that graph-based retrieval and dense retrieval capture complementary aspects of relevance. Specifically, graph retrieval exploits structured relations among biomedical entities to propagate query relevance across graph, while dense retrieval focuses on semantic similarity between the query and textual documents. By combining these two paradigms, the system can leverage both global structural knowledge and fine-grained semantic information.  Given a medical question $Q$, we decompose it into a set of sub-questions for retrieval. Specifically, we leverage the Large Language Model ($\phi_{\text{LLM}}$) to perform this decomposition. Instructed by a tailored prompt, $\phi_{\text{LLM}}$ breaks down the multi-hop question $Q$ into $m$ distinct, simpler sub-questions. This decomposition helps isolate different pieces of information required to answer the original question. For each sub-question $q$, graph-based retrieval and dense vector retrieval are performed in parallel. To be concrete, a one-shot prompt is first used to extract a set of query entities, denoted as $\mathcal{C}_{q} = \{ c_1, c_2, \ldots, c_k \}$. Each query entity $c_i$ is encoded into a dense vector $\mathbf{c}_i$ using an embedding model (\textit{i.e.}, Contriever-MSMARCO). We then align each $c_i$ to the most similar entity $e \in \mathcal{E}$ in the previously constructed KG $G$ based on cosine similarity between embeddings. Accordingly, the aligned entities for sub-question $q$ are defined as $\mathcal{E}_q = \{ e_1, e_2, \ldots, e_k \}$ with
\begin{equation}
e_i = \operatorname*{arg\,max}_{e \in \mathcal{E}} 
\frac{\mathbf{c}_i \cdot \mathbf{e}}{\|\mathbf{c}_i\| \, \|\mathbf{e}\|},
\end{equation}
where $\mathbf{e}$ denotes the embedding of $e$. We refer to this embedding-based alignment between query entities and graph entities as entity linking, which allows flexible matching between query mentions and KG entities without relying on predefined ontology mappings. This step effectively bridges natural language queries and structured knowledge representations, enabling the graph retriever to operate on semantically aligned entities.

Subsequently, we employ a graph foundation model-based retriever GFM-RAG-8M \cite{luo2025gfm} to propagate query relevance via message passing. By leveraging the capability of GNNs to model structured dependencies, the retriever can propagate relevance signals from the aligned entities to their neighboring entities and associated documents in the KG. This process allows the system to capture indirect or multi-hop relations among biomedical concepts, which are common in clinical knowledge bases \cite{gnn-rag,RGNN-Ret}. As a result, we obtain a normalized relevance score distribution over entities and documents in the KG. The relevance scores of entity nodes are then mapped to documents using the entity-to-document inverted index $\mathbf{M}$. Finally, the graph retrieval score for each document (\textit{i.e.}, $S_{\text{graph}}(d_i)$), is computed by aggregating the relevance probabilities of its associated entities. Meanwhile, the sub-question $q$ is encoded into a dense vector $\mathbf{q}=\phi(q)$. Afterwards, we compute the cosine similarity between the query vector and each document vector to obtain the dense retrieval score:
\begin{equation}
S_{\text{vec}}(d_i) = \frac{\mathbf{q} \cdot \mathbf{d}_{i}}{\|\mathbf{q}\| \, \|\mathbf{d}_{i}\|}.
\end{equation}
To jointly exploit structural information from graph-based retrieval and semantic relevance from dense retrieval, we adopt Reciprocal Rank Fusion (RRF) to aggregate the results of the two paths. RRF is adopted as it provides a robust rank-based fusion strategy that does not require score normalization across heterogeneous retrieval models. The final relevance score of document $d_i$ is computed as
\begin{equation}
\label{fuse_score}
S_{\text{final}}(d_i) = \frac{1}{\eta + \text{rank}_{\text{graph}}(d_i)} + \frac{1}{\eta + \text{rank}_{\text{vec}}(d_i)},
\end{equation}
where $\text{rank}_{\text{graph}}(d_i)$ and $\text{rank}_{\text{vec}}(d_i)$ denote the ranking positions of document $d_i$ in the ranked lists produced by graph-based retrieval and dense retrieval, respectively, and $\eta$ is a smoothing constant. Documents are ranked according to $S_{\text{final}}$, and the top-$k$ documents are returned as initial evidence for subsequent reasoning.

\subsection{Iterative Retrieval-reasoning for Question Answering}
\label{IterativeQA}

For complex medical QA, relying on a single retrieval step is often insufficient, as the system may fail to gather all relevant information at once. Moreover, retrieval and reasoning are inherently complementary: intermediate reasoning results can reveal missing knowledge and guide subsequent retrieval, while newly retrieved evidence further supports deeper reasoning. Therefore, an iterative retrieval–reasoning process is necessary to gradually accumulate evidence and refine the reasoning path. To address this limitation, we embed dual-path retrieval within an LLM-driven iterative retrieval-reasoning framework, which enables the knowledge acquired in previous rounds to explicitly guide subsequent retrieval and reasoning steps. Specifically, at the $t$-th reasoning round, the model maintains a history state $\mathcal{H}_t = \{ Q, \mathcal{X}_1, \mathcal{X}_2,...,\mathcal{X}_{t-1}\}$, which consists of the original question $Q$ and all previously generated sub-question–answer pairs $\{ \mathcal{X}_1, \mathcal{X}_2,...,\mathcal{X}_{t-1}\}$. Each pair is defined as
$\mathcal{X}_{i} = \{ (q_{i,j}, a_{i,j}) \mid j = 1, \dots, m \}$, where $q_{i,j}$ denotes the $j$-th sub-question generated at the $i$-th reasoning round and $a_{i,j}$ represents the corresponding sub-answer produced for $q_{i,j}$. Based on $\mathcal{H}_t$, the LLM analyzes the current reasoning context and identifies remaining knowledge gaps before producing a new set of sub-questions $\{ q_{t,1}, q_{t,2}, \dots, q_{t,m} \}$. 
These sub-questions aim to explore unexplored aspects of the problem and retrieve additional evidence required for answering the original query. Crucially, the initial question decomposition described in Section 3.2 is fundamentally a unified special case of this iterative mechanism at the first round ($t=1$), where the history state is empty (i.e., $\mathcal{H}_1 = \{Q\}$). In this bootstrap step, the decomposition focuses on breaking down the global structure of $Q$. In all subsequent rounds ($t>1$), as $\mathcal{H}_t$ accumulates evidence, the sub-question generation dynamically shifts to a gap-filling strategy based on the identified missing knowledge. For each sub-question $q_{t,j}$, we perform the dual-path retrieval process described in Section \ref{3.2} to collect relevant evidence and generate a corresponding sub-answer $a_{t,j}$. The newly obtained sub-question–answer pairs are then incorporated into the history state to form $\mathcal{H}_{t+1}$, which serves as the context for the next reasoning round. Through this iterative process, the system gradually expands its evidence pool and refines the reasoning trajectory. The reasoning process continues until a predefined maximum number of reasoning rounds $T$ is reached. Finally, the LLM integrates the final history state $\mathcal{H}_T$ to generate the answer to the original question. The overall framework of the proposed Hybrid-IR approach is summarized in Algorithm~\ref{alg:hybrid_ir}.

\begin{algorithm}[t]
    \caption{Proposed Hybrid-IR Framework}
    \label{alg:hybrid_ir}
    \begin{algorithmic}[1] 
        \Require Medical question $Q$; LLM $\phi_{\text{LLM}}$; graph index $G$; vector index ${V}$; medical corpus $\mathcal{D}$; maximum number of reasoning rounds $T$; number of sub-questions generated at each reasoning round $m$.
        \Ensure Final answer.
        
        \State Initialize the reasoning history $\mathcal{H}_1 \leftarrow \{Q\}$
        \For{$t = 1, 2, \ldots, T$}
            \State Generate sub-questions $\{q_{t,1},\ldots,q_{t,m}\}$ using $\phi_{\text{LLM}}$ given $\mathcal{H}_t$.
            
            \For{$j = 1, 2, \ldots, m$}
                \State Calculate graph retrieval scores $S_{\text{graph}}$ based on GFM-RAG-8M.
                \State Calculate dense retrieval scores $S_{\text{vec}}$ via the dense retriever.
                \State Obtain the final scores $S_{\text{final}}$ based on Eq.~\eqref{fuse_score}.
                \State Retrieve top-$k$ documents according to $S_{\text{final}}$.
                
                \State Generate sub-answer $a_{t,j}$ based on the sub-question $q_{t,j}$ and the retrieved documents.
                \State Update the history $\mathcal{H}_{t} \leftarrow \mathcal{H}_t \cup \{(q_{t,j}, a_{t,j})\}$.
            \EndFor
            \State Set $\mathcal{H}_{t+1} \leftarrow \mathcal{H}_t$.
        \EndFor
        \State Generate the final answer based on $\mathcal{H}_{T}$ and $\phi_{\text{LLM}}$.
        \State \Return the final answer. 
    \end{algorithmic}
\end{algorithm}

\section{Experiments}

In this section, we conduct extensive experiments to evaluate the effectiveness of the proposed Hybrid-IR framework for medical QA. We compare Hybrid-IR with a range of strong retrieval-augmented baseline methods on multiple medical QA benchmarks, and further analyze the contribution of each component through ablation studies and case analysis.

\subsection{Experimental Setup}

\subsubsection{Dataset} We conducted experiments on three widely used medical QA datasets, all of which contain complex biomedical questions that often require multi-step reasoning across multiple pieces of medical knowledge. Specifically, MedQA \cite{jin2021disease} consists of USMLE-style questions selected by medical examination experts from multiple medical question banks, many of which involve clinically complex scenarios that require integrating diverse medical concepts. MedMCQA \cite{pal2022medmcqa} contains medical questions from two Indian examination bodies. MMLU-Med \cite{hendrycks2020measuring} is a subset of the MMLU \cite{hendrycks2020measuring} dataset, which includes  questions from six biomedical subjects, namely clinical knowledge, medical genetics, anatomy, specialty medicine, college biology, and college medicine. All these datasets contain multiple-choice questions with four answer options and cover biomedical knowledge ranging from high school to professional medical levels. Following previous works \cite{sohn2024rag2,xiong2024benchmarking,xiong2024improving}, we evaluate our method on the full development set of each dataset. For retrieval, we use the Textbooks corpus \cite{jin2021disease}, which comprises 18 authoritative medical textbooks commonly used as essential reference materials for the United States Medical Licensing Examination (USMLE) \cite{usmle2026}.

\subsubsection{Evaluation Protocol} Our experiments are organized into two parts, \textit{i.e.}, baseline methods without retrieval, which evaluates the intrinsic zero-shot reasoning capabilities of LLMs without access to external corpora, and retrieval-augmented baseline methods, where external knowledge is retrieved to assist the answer generation process. In the non-retrieval setting, we evaluated the zero-shot reasoning capabilities of the open-source LLM Meta-Llama-3-8B-Instruct \cite{llama3modelcard} and the commercial LLM GPT-4o-mini \cite{openai2024hellogpt4o}. Meta-Llama-3-8B-Instruct represents a state-of-the-art open-source model at a comparable scale, and is referred to as Llama-3-8B for simplicity. GPT-4o-mini is a cost-effective variant of the latest closed-source LLM produced by OpenAI, which exhibits performance comparable to GPT-4 across multiple benchmarks. In the retrieval-augmented setting, these two models are used as backbone LLMs and are integrated with RAG-based methods to evaluate whether the proposed Hybrid-IR framework can further enhance performance without task-specific fine-tuning. Following prior studies on medical QA, we report the answer accuracy as the primary evaluation metric, which is defined as the proportion of questions for which the model selects the correct option among the four candidates.

\subsubsection{Baseline Methods} To validate the effectiveness of Hybrid-IR, we compare it with a set of representative baseline methods. All comparative experiments are performed using the two aforementioned backbone LLMs under the same evaluation protocol. MedCPT \cite{jin2023medcpt} is a retriever pre-trained using user search logs from PubMed. MedRAG \cite{xiong2024benchmarking} employs a hybrid retrieval strategy that combines multiple sparse and dense retrievers to collect candidate evidence from the MedCorp corpus, followed by reranking using Reciprocal Rank Fusion (RRF) \cite{rrf}. HippoRAG \cite{jimenez2024hipporag} leverages a hippocampus-inspired KG to enable multi-hop reasoning over entity relationships. GFM-RAG \cite{luo2025gfm} incorporates graph-based message passing over a KG to model dependencies among entities and performs structured evidence retrieval. MedGraphRAG \cite{wu2025medical} constructs a medical KG to retrieve structured evidence from biomedical literature. i-MedRAG \cite{xiong2024improving} introduces an iterative retrieval strategy that progressively refines retrieved evidence across multiple reasoning steps. 

\subsubsection{Implementation Details} In the KG-index construction phase, we use Llama-3-8B with the OpenIE prompts described in HippoRAG \cite{jimenez2024hipporag} to extract entities and relations from unstructured medical texts. Entity representations are obtained using Contriever-MSMARCO \cite{izacard2021contriever}, and entity resolution is performed based on cosine similarity with a threshold of 0.85. This threshold is selected empirically to balance the preservation of meaningful semantic relations and the suppression of noisy entity associations. During online retrieval, we employ GFM-RAG-8M \cite{luo2025gfm} as the graph encoder for message passing and relevance estimation. For vector retrieval, MedCPT is adopted as the text embedding model to capture dense semantic information. To ensure fairness, all methods are evaluated using the same retrieval corpus (\textit{i.e.}, Textbooks \cite{jin2021disease}). For each baseline, we strictly follow the implementation settings and hyperparameters reported in the corresponding paper. The maximum number of reasoning rounds is set to 3 and the number of generated sub-questions per round is set to 3.

\subsection{Experimental Results}

The comparative performance of Hybrid-IR against various baseline methods in three medical QA datasets is presented in Table~\ref{tab:main}, where the best and second-best results are highlighted in bold and underlined, respectively. In general, Hybrid-IR outperforms all baselines under both Llama-3-8B and GPT-4o-mini backbones. Concretely, under the Llama-3-8B-Instruct backbone, Hybrid-IR outperforms the strongest baseline, \textit{i.e.}, i-MedRAG (67.2\%), by a margin of 2.1 points on average, and substantially surpasses single-modality retrieval methods, such as GFM-RAG (61.3\%) and MedRAG (60.7\%). This improvement demonstrates that Hybrid-IR effectively compensates for the limited parametric medical knowledge of smaller LLMs by integrating external structured and unstructured evidence through an iterative retrieval-reasoning loop guided by sub-question decomposition.

When applied to GPT-4o-mini, which already achieves a strong baseline accuracy, Hybrid-IR still delivers improvement. While strong backbone models may be less sensitive to retrieval augmentation and can be affected by irrelevant evidence, Hybrid-IR achieves further gains by improving evidence aggregation. Overall, the results confirm that the proposed iterative dual-path retrieval strategy provides reliable performance improvements across various datasets and backbone models. These improvements can be attributed to two key factors. 
First, the dual-path retrieval mechanism enables Hybrid-IR to capture complementary evidence from both structured KGs and unstructured medical texts, which provides a comprehensive evidence pool for complex medical reasoning tasks. 
Second, the reasoning–retrieval loop allows intermediate reasoning results to generate targeted sub-questions that guide subsequent retrieval, thereby progressively refining the retrieved evidence and reducing the impact of irrelevant information.

\begin{table}[!t]
    \centering
    \caption{Experimental results on three medical QA datasets. All results are reported in accuracy (\%).}
    \label{tab:main}
    \setlength{\tabcolsep}{6pt}
    \resizebox{\columnwidth}{!}{%
        \begin{tabular}{lcccc}
            \toprule
            Method & MedQA & MedMCQA & MMLU-Med & Average \\
            \midrule
            Llama-3-8B-Instruct & 57.7 & 53.5 & 69.5 & 60.2 \\
            \midrule
            \hspace{1em} +MedCPT($k=1$) & 55.3 & 51.3 & 65.8 & 57.5 \\
            \hspace{1em} +MedRAG & 56.4 & 56.6 & 69.2 & 60.7 \\
            \hspace{1em} +i-MedRAG & \underline{71.8} & 57.4 & \underline{72.3} & \underline{67.2} \\
            \hspace{1em} +GFM-RAG & 60.3 & \underline{58.8} & 64.8 & 61.3 \\
            \hspace{1em} +HippoRAG & 57.9 & 56.4 & 62.1 & 58.8 \\
            \hspace{1em} +MedGraphRAG & 57.8 & 54.3 & 65.2 & 59.1 \\
            \hspace{1em} +\textbf{Ours} & \textbf{72.2} & \textbf{59.8} & \textbf{75.8} & \textbf{69.3} \\
            \midrule
            GPT-4o-mini & 80.3 & 68.5 & \textbf{88.9} & \underline{79.2} \\
            \midrule
            \hspace{1em} +MedCPT($k=1$) & 74.3 & 65.1 & 84.9 & 74.8 \\
            \hspace{1em} +MedRAG & 75.5 & 65.6 & 85.1 & 75.4 \\
            \hspace{1em} +i-MedRAG & \underline{80.5} & \underline{68.2} & 85.9 & 78.2 \\
            \hspace{1em} +GFM-RAG & 78.5 & 66.2 & 85.9 & 76.9 \\
            \hspace{1em} +HippoRAG & 75.1 & 64.3 & 85.3 & 74.9 \\
            \hspace{1em} +MedGraphRAG & 71.5 & 62.8 & 83.4 & 72.6 \\
            \hspace{1em} +\textbf{Ours} & \textbf{82.9} & \textbf{68.6} & \underline{86.8} & \textbf{79.4} \\
            \bottomrule
        \end{tabular}
    }
\end{table}

\subsection{Ablation Study}

\begin{table}[!t]
    \centering
    \caption{Ablation study using GPT-4o-mini as the backbone model. All results are reported in accuracy (\%).}
    \label{tab:ablation_study}
    \setlength{\tabcolsep}{6pt}
    \resizebox{\columnwidth}{!}{%
        \begin{tabular}{ccc|ccc}
            \toprule
            Graph & Vector & Iterative & MedQA & MedMCQA & MMLU-Med \\
            \midrule
            $\checkmark$ & - & - & 78.5 & 66.2 & 85.9 \\
            - & $\checkmark$ & - & 75.5 & 65.6 & 85.1 \\
            $\checkmark$ & $\checkmark$ & - & 78.7 & 66.6 & 84.8 \\
            $\checkmark$ & - & $\checkmark$ & 76.4 & \underline{68.3} & \underline{86.5} \\
            - & $\checkmark$ & $\checkmark$ & \underline{80.5} & 68.2 & 85.9 \\
            $\checkmark$ & $\checkmark$ & $\checkmark$ & \textbf{82.9} & \textbf{68.6} & \textbf{86.8} \\
            \bottomrule
        \end{tabular}%
    }
\end{table}

Here, we perform an ablation study using GPT-4o-mini as the backbone model to evaluate the contribution of each component in Hybrid-IR. The results in Table~\ref{tab:ablation_study} reveal several important observations. First, utilizing either graph-based retrieval or vector-based retrieval alone yields moderate performance (\textit{i.e.}, 78.5\% and 75.5\% on MedQA, respectively). This indicates that each retrieval paradigm captures different aspects of the evidence required for medical QA. Second, simply combining graph and vector retrieval without iteration leads to only marginal improvements (\textit{i.e.}, 78.7\% vs 78.5\% on MedQA), which suggests that parallel retrieval alone is insufficient to fully exploit the complementary information provided by the two retrieval paths. Third, introducing iterative reasoning to a single retrieval path brings limited gains, as the reasoning process remains constrained by the restricted evidence source of that retrieval strategy. For example, the vector-based iterative setting achieves 80.5\% on MedQA, which is still lower than the full model. Finally, when graph retrieval, vector retrieval, and iterative reasoning are jointly integrated, the model achieves the best performance across all datasets. This demonstrates that Hybrid-IR benefits from a reasoning–retrieval loop in which intermediate reasoning results guide subsequent retrieval, which enables the model to progressively accumulate complementary evidence from both structured KGs and unstructured texts. This result highlights the importance of coupling retrieval and reasoning within a unified reasoning–retrieval loop for complex medical QA. It is worth noting that, unlike graph retrieval, dense retrieval, and iterative reasoning, question decomposition is not ablated as a standalone component. As described in Section 3.3, sub-question generation is performed within each reasoning round rather than as a one-time preprocessing step. In the first round, they are derived from the original question to expose its global reasoning structure. In later rounds, they are dynamically generated according to the accumulated reasoning history $\mathcal{H}_t$ to fill unresolved evidence gaps. Therefore, question decomposition is an integral part of the iterative retrieval-reasoning process, and its effect is reflected in the iterative variants reported in Table~\ref{tab:ablation_study}.

\subsection{Parametric Sensitivity}

\begin{figure}[!t]
\centerline{\includegraphics[width=0.6\linewidth]{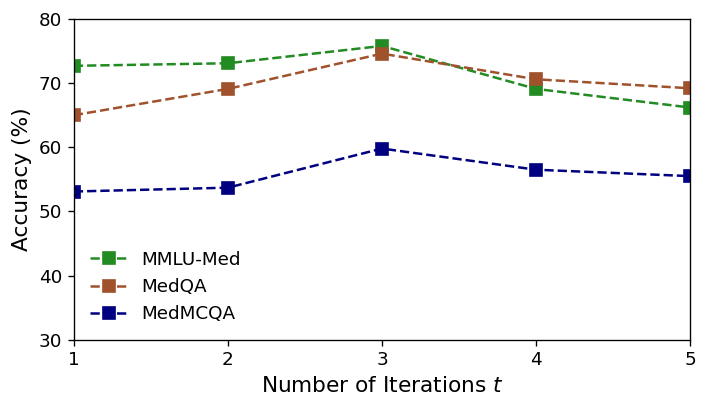}}
\caption{Sensitivity analysis of the number of iterations $t$ on different datasets.}
\label{fig:performance_chart}
\end{figure}

We analyze the sensitivity of Hybrid-IR to the number of iterations $t$ using the Llama-3-8B backbone, as shown in Fig.~\ref{fig:performance_chart}. Overall, the performance of Hybrid-IR remains relatively stable within a reasonable range of $t$, suggesting that Hybrid-IR is not highly sensitive to this hyperparameter. As $t$ increases from 1 to 3, the model can be consistently improved, which indicates that iterative retrieval effectively facilitates the integration of multi-step evidence. When $t>3$, further increasing the number of iterations brings no clear improvement and may slightly degrade performance. In terms of computational cost, the overhead of Hybrid-IR increases approximately linearly with the number of iterations, since each additional round introduces an extra retrieval-reasoning cycle. As shown in Fig.~\ref{fig:performance_chart}, the performance improves from $t=1$ to $t=3$ and then tends to saturate, while larger values of $t$ bring no consistent benefit but incur additional computation. Therefore, we set $t=3$ as the default value, which provides a practical balance between accuracy and efficiency.

\subsection{Case Study}

We present a representative case from MedQA to illustrate the multi-step reasoning capability of Hybrid-IR, as shown in Table~\ref{tab:case_study}. Although Table~\ref{tab:case_study} presents two representative sub-questions for illustration, the model decomposes the original question into multiple intermediate sub-questions during the reasoning process. In the first iteration, the model identified the pathogen as the rabies virus. Vector-based retrieval provides evidence at the symptom-level (\textit{e.g.}, agitation), while graph-based retrieval captures an association chain linking travel history related to cave exploration with bats and the rabies virus. 
Based on this intermediate reasoning result, the system initiates a second retrieval step to further investigate the underlying infection mechanism. In this iteration, the KG retrieves the key relation indicating retrograde axoplasmic transport, which directly leads to the correct answer. This example demonstrates how Hybrid-IR decomposes a complex medical question into tractable sub-questions and progressively integrates complementary evidence through a reasoning–retrieval loop, where intermediate reasoning results guide subsequent dual-path retrieval over both structured KGs and unstructured texts.

\begin{table}[!t]
    \centering
    \caption{A case study from MedQA illustrating the iterative dual-path reasoning process of Hybrid-IR.}

    \label{tab:case_study}
    
    \scriptsize 
    
    \setlength{\tabcolsep}{1pt}
    
    \renewcommand{\arraystretch}{1.3}
    
    \begin{tabularx}{\linewidth}{|c|>{\raggedright\arraybackslash}X|>{\raggedright\arraybackslash}X|}
        \hline
        
        \multicolumn{3}{|c|}{\textbf{Question}} \\
        \hline
        \multicolumn{3}{|p{\dimexpr\linewidth-2\tabcolsep-2\arrayrulewidth}|}{%
            A man who had traveled to Mexico for cave exploration was hospitalized due to agitation, photophobia... Through which route did the virus spread to the CNS?
        } \\
        \hline
        
        & \multicolumn{2}{c|}{\textbf{Iterative Reasoning}} \\
        \hline
        \textbf{} & \textbf{Iteration 1} & \textbf{Iteration 2} \\
        \hline
        
        \textbf{Sub-Q} & 
        Which virus causes agitation and photophobia after cave exploration?& 
        What is the route by which the rabies virus enters the CNS? \\
        \hline
        
        \textbf{Vector} & 
        Progression of \textbf{rabies virus}: agitation, photophobia, hydrophobia ... & 
        Effective neuroinvasion by the rabies virus is a prerequisite for CNS infection.\\
        \hline
        
        \textbf{Graph} & 
        (caves, habitat, bats) \par
        (bats, carry, \textbf{rabies virus}) \par
        (\textbf{rabies virus}, causes, photophobia) & 
        (rabies virus, spreads via, \textbf{retrograde axonal transport}) \par
        (rabies virus, destination, CNS) \\
        \hline
        
        \multicolumn{3}{|c|}{\textbf{Answer:} Retrograde migration up peripheral nerve axons} \\
        \hline
        
    \end{tabularx}
\end{table}

\section{Conclusion}

In this paper, we present a novel framework termed ``Hybrid-IR'' designed to enhance the performance of LLMs for complex medical QA, which integrates complementary dual-path retrieval to jointly capture fine-grained semantic information from unstructured texts and structured contextual knowledge from KGs. Moreover, the reasoning–retrieval loop allows intermediate reasoning results to generate sub-questions that guide subsequent retrieval, which enables the model to progressively refine the retrieved evidence. While we have demonstrated the effectiveness of Hybrid-IR on standard multiple-choice medical QA benchmarks, exploring its capabilities in open-ended clinical inquiries and real-world clinical decision-support scenarios remains an important direction for our future work.

\begin{credits}
\subsubsection{\ackname} This work was supported in part by the National Natural Science Foundation of China (No. 62506171), the Natural Science Foundation of Jiangsu Province (No. BK20241469), and the Fundamental Research Funds for the Central Universities (No. YDZX2026052).
\end{credits}

%
%
%
\bibliographystyle{splncs04}
\bibliography{references}
%




\end{document}